\documentclass[10pt,twocolumn,letterpaper]{article}

\usepackage{cvpr}
\usepackage{times}
\usepackage{epsfig}
\usepackage{graphicx}
\usepackage{amsmath}
\usepackage{amssymb}
\usepackage{subcaption}
\usepackage{bm}
\usepackage{enumitem}
\usepackage{booktabs}
\usepackage{algpseudocode,algorithm,algorithmicx}  
\usepackage{color}
\usepackage[table]{xcolor}
\definecolor{col1}{RGB}{234,209,220}
\definecolor{col2}{RGB}{217, 211, 0}
\definecolor{darkblue}{rgb}{0.0, 0.0, 1.0}

\newcommand{\ea}{\textit{et al}.}

\usepackage[pagebackref=true,breaklinks=true,letterpaper=true,colorlinks,bookmarks=false]{hyperref}

\cvprfinalcopy 

\def\cvprPaperID{****} 

\ifcvprfinal\fi
\begin{document}

\title{Unsupervised Monocular Depth and Ego-motion Learning with \\Structure and Semantics}

\author{Vincent Casser$^{1,2}$         \and
        Soeren Pirk$^1$ \and 
        Reza Mahjourian$^{1,3}$ \and
        Anelia Angelova$^1$
}

\author{Vincent Casser$^{1,2}$\thanks{Work done while at Google Brain.} \\
{\tt\small casser@google.com}
\and
Soeren Pirk$^1$\\
{\tt\small pirk@google.com} \\
\and
Reza Mahjourian$^{1,3}$\\
{\tt\small rezama@google.com} \\
\and
Anelia Angelova$^{1}$\\
{\tt\small anelia@google.com}\\
\vspace{0.1cm}
 $^1$ Robotics at Google, Google Brain $^2$ Harvard University $^3$ University of Texas at Austin
}
\maketitle

\begin{abstract}
We present an approach which takes advantage of both structure and semantics for unsupervised monocular learning of depth and ego-motion. More specifically, we model the motion of individual objects and learn their 3D motion vector jointly with depth and ego-motion. We obtain more accurate results, especially for challenging dynamic scenes not addressed by previous approaches. This is an extended version of Casser et al.~\cite{casser2019depth}. Code and models have been open sourced at: 
\textbf{https://sites.google.com/corp/view/struct2depth}.
\end{abstract}

\section{Introduction}

\noindent Predicting scene depth  and agent ego-motion from input imagery is important for robot navigation, both for indoors and outdoors settings.  
While supervised dense depth prediction has been successful ~\cite{eigen2014depth}, we here consider joint learning of depth and ego-motion from monocular input videos only. Unsupervised monocular or stereo-based learning has also shown progress recently~\cite{zhou2017unsupervised,godard2017monodepth}, but prior work has not been successful at dynamic scenes.

\begin{figure}[t]
    \centering        
    \includegraphics[width=1.0\linewidth]{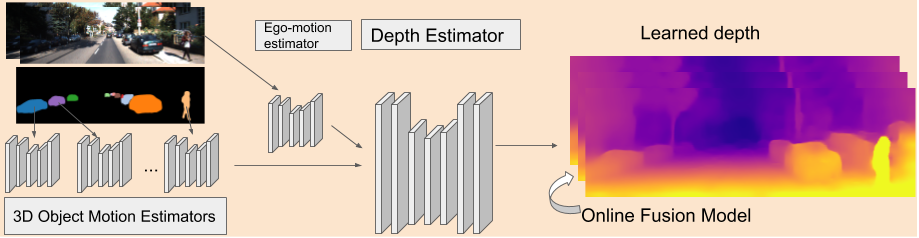}
    \caption{Our method utilizes 3D geometry structure and semantics during learning by modeling motions of individual objects, ego-motion and scene depth in a principled way. Furthermore, a refinement approach adapts the model on the fly in an online fashion.}
    \label{fig:approach}
\end{figure}

We present an approach that explicitly {\it models 3D motions} of moving objects, together with camera ego-motion and scene depth, and adapts to new environments by learning with an {\it online refinement} of multiple frames (Figure~\ref{fig:approach}).
With the newly introduced motion handling and the proposed object size constraint, this approach is the first to effectively learn from highly dynamic scenes in a monocular setting. Our approach introduces both structure and semantics in the learning process by representing objects in 3D and modeling motion as SE3 transforms; this is trained from uncalibrated monocular videos in a fully differentiable manner. 
We further introduce an online refinement method for domain transfer in this unsupervised learning setting, which can be applied independently of the base method.
This work is an extended version of~\cite{casser2019depth}. We here present additionally new results on the challenging Cityscapes dataset with prevalent dynamic scenes and on ego-motion.
Our algorithm yields significant improvements on two publicly available datasets and on both depth and ego-motion estimation, compared to the state-of-the-art, especially on dynamic scenes.
Furthermore, we evaluate direct domain transfer, by training on one dataset and testing on another, without fine-tuning.

\textbf{Setup:}
The main learning setup is unsupervised learning of depth and ego-motion from monocular video~\cite{zhou2017unsupervised}, where the only source of supervision is obtained from the video itself. No depth sensor supervision is used. Objects' masks are introduced from an off-the-shelf algorithm during training only. During inference, only a still input image is needed to predict depth, and two images to predict ego-motion. \textbf{Runtime:} our model runs at 50 FPS and 30 FPS on a NVIDIA GeForce 1080Ti for batch 4 and 1, respectively.

\begin{figure*}[t]
    \begin{center}
    \includegraphics[width=0.98\linewidth]{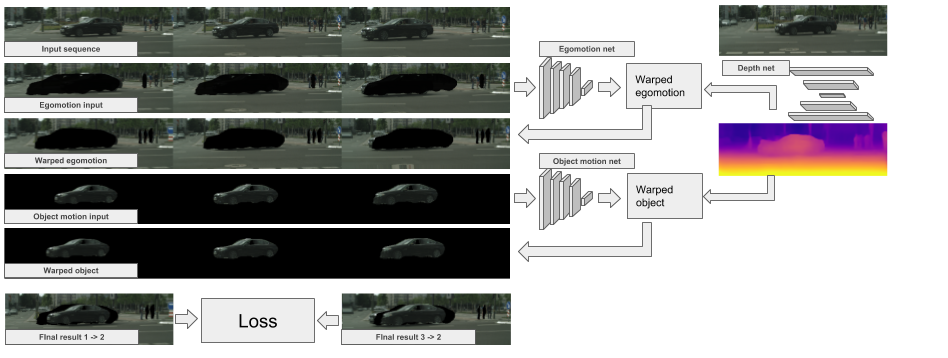}
    \end{center}
    \caption{Schematic of the warping sequence for our method: first object masks are used to remove regions with movement; then object ego-motion is computed; after that individual object motion is computed, but this is done using the output of the image warped according to ego-motion. The final warped images both previous and next (with a validity mask) are compared in RGB space to the original image.}
    \label{fig:approach_detailed}
\end{figure*}

\section{Previous Work}
Recent methods have demonstrated supervised learning of scene depth from input imagery \cite{eigen2014depth,liu2015learning,laina2016deeper,mahjourian2017geometry}. Depth information provided by a sensor, such as a LiDAR, is used as supervision.
In parallel to supervised learning techniques, unsupervised image-to-depth learning has been proposed ~\cite{zhou2017unsupervised,garg2016unsupervised,godard2017monodepth,ummenhofer2017demon,mahjourian2018unsupervised}, where the only supervision is obtained from a monocular video.
The work of Garg et al.~\cite{garg2016unsupervised} introduced joint learning of depth and ego-motion in a neural based framework. Zhou et al.~\cite{zhou2017unsupervised} proposed the first fully differentiable  deep neural network approach for unsupervised learning of depth and ego-motion, and showed it outperforms prior approaches which used depth sensors as supervision.
Many subsequent works have further improved the quality of depth and ego-motion~\cite{Yang2017unsupervised,ummenhofer2017demon,mahjourian2018unsupervised,pilzer2018unsupervised,yang2018lego,yin2018geonet,wang2018learning,Kuznietsov2017semisupervised}. Some of these approaches have successfully used stereo pair videos during training~\cite{godard2017monodepth,ummenhofer2017demon,zhan2018unsupervised,yang2018every,godard2018digging} to also produce a single image-based depth estimation. These methods tend to achieve better quality results, due to the extra camera input.

\section{Depth and Ego-motion Learning}
We here present an approach which is able to model dynamic scenes by modeling object motion, and that can optionally adapt its learning strategy with an online refinement technique. Note that both ideas are tangential and can be used either separately or jointly. See ~\cite{casser2019depth} for details. During training, the method operates on sequences of three consecutive RGB images\footnote{While in theory the whole formulation can be done with two consecutive frames, for convenience, we consider three images in a sequence and impose constraints between two pairs of frames.} $(I_1, I_2, I_3)\in \mathbb{R}^{H\times W\times 3}$, and camera intrinsics matrix $K\in \mathbb{R}^{3\times 3}$.
Depth is predicted by learning a depth function $\theta:\mathbb{R}^{H\times W\times 3}\rightarrow \mathbb{R}^{H\times W}$, which is a fully convolutional encoder-decoder neural network producing a dense depth map $D_i=\theta(I_i)$ from a single RGB frame. Ego-motion is predicted by an ego-motion network $\psi_E:\mathbb{R}^{2\times H\times W\times 3}\rightarrow \mathbb{R}^{6}$ which takes a sequence of two RGB images as input and produces the SE3 transform between the frames, i.e. 6-dimensional transformation vector $E_{1\rightarrow 2}=\psi_E(I_1, I_2)$ specifying translation and rotation parameters between the frames. 

Let us suppose that the depth network output is providing an adequate depth of the scene per frame, then using it, we can represent points in 3D space. Further, given the ego-motion between consecutive frames, we can transform the scene and project it to obtain the neighbouring frame.
More specifically, by using a differentiable image warping operator $\phi(I_i, D_j, E_{i\rightarrow j})\rightarrow \hat{I}_{i\rightarrow j}$, we can inverse-warp any source RGB-image $I_i$ into $I_j$ given corresponding depth estimate $D_j$ and an ego-motion estimate $E_{i\rightarrow j}$. Here, $\phi$ performs the warping by reading from transformed image pixel coordinates, setting $\hat{I}_{i\rightarrow j}^{xy} = I_i^{\hat{x}\hat{y}}$, where $[\hat{x},\hat{y},1]^T=KE_{i\rightarrow j} (D_j^{xy} \cdot K^{-1} [x,y,1]^T)$.
The latter construct succinctly denotes projecting the depth into a 3D point cloud, then transforming it according to $E_{i\rightarrow j}$ and then projecting the transformed 3D point cloud into image space.
In practice, we always warp the outer images towards the center frame within a sequence.
The supervisory signal is then established using a photometric loss comparing the projected scene onto the next frame $\hat{I}_{i\rightarrow j}$ with the actual next frame $I_j$ image in RGB space, for example using a reconstruction loss: $L_{rec}=\min(\Vert \hat{I}_{i\rightarrow j}-I_j\Vert$.

\subsection{Motion Model}
In order to handle highly dynamic scenes, we model motions of individual objects.
Namely, we introduce a third component $\psi_M$ to the model, which is specialized to predicting motion of objects in 3D (Figure~\ref{fig:approach}).
It uses the same network structure as the ego-motion network $\psi_E$ but trains to separate weights. The object motion model takes an RGB image sequence as input, complemented by pre-computed instance segmentation masks. The motion model learns to predict the transformation vectors per each object in 3D-space, which create the observed object appearance in the respective target frame when applied to the camera. In the new model, the final warping result is a combination of the individual warping from moving objects, and the ego-motion. (Figure~\ref{fig:approach_detailed}).
In order to compute ego-motion, object motions are masked out of the images first, i.e. the static scene binary mask is applied to all images in the sequence by element-wise multiplication, before feeding the sequence to the ego-motion model.
The static background is generated by a single warp based on $\psi_E$, whereas all segmented objects are then added by their appearance being warped first according to $\psi_E$ and then $\psi_M$. Outlines of potentially moving objects are provided by an off-the-shelf algorithm \cite{he2017mask} and are needed only for training (similar to prior work that use optical flow~\cite{yang2018every} that is not trained on either of the datasets of interest). Our approach not only models objects in 3D but also learns their motion on the fly, together with scene depth and ego-motion.

\begin{figure*}[t]
    \centering
    \includegraphics[width=0.24\linewidth]{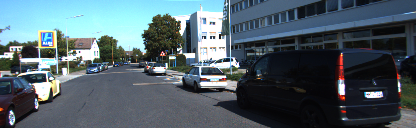}
    \includegraphics[width=0.24\linewidth]{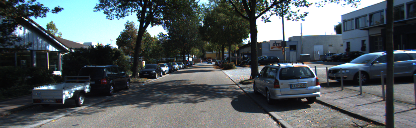}
    \includegraphics[width=0.24\linewidth]{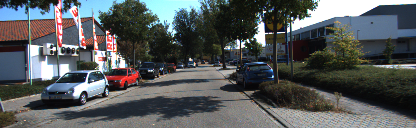}
    \includegraphics[width=0.24\linewidth]{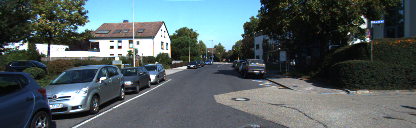}
    \includegraphics[width=0.24\linewidth]{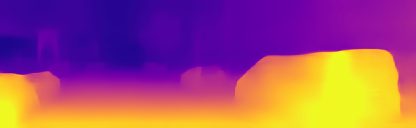}
    \includegraphics[width=0.24\linewidth]{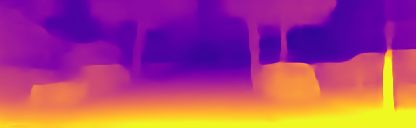}
    \includegraphics[width=0.24\linewidth]{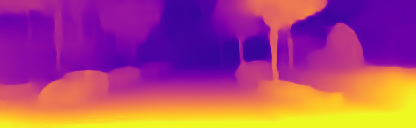}
    \includegraphics[width=0.24\linewidth]{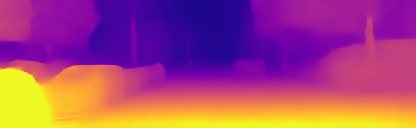}
    \includegraphics[width=0.24\linewidth]{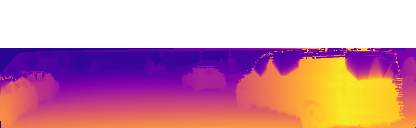}
    \includegraphics[width=0.24\linewidth]{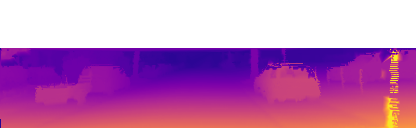}
    \includegraphics[width=0.24\linewidth]{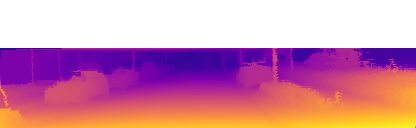}
    \includegraphics[width=0.24\linewidth]{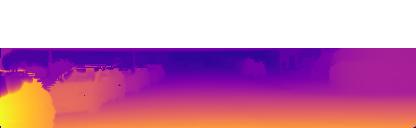}
    \caption{Example results of depth estimation. Each column shows an input image, depth prediction of our method, and ground truth depth. KITTI dataset.}
    \label{fig:main_kitti}
\end{figure*}

\begin{table*} [h!]
  \centering
  \resizebox{1.0\textwidth}{!}{
  \begin{tabular}{|l|c|c|c||c|c|c|c|c|c|c|}
  \hline
  Method & Supervision? & Motion? & Cap & \cellcolor{col1}Abs Rel & \cellcolor{col1}Sq Rel & \cellcolor{col1}RMSE  & \cellcolor{col1}RMSE log & \cellcolor{col2}$\delta < 1.25 $ & \cellcolor{col2}$\delta < 1.25^{2}$ & \cellcolor{col2}$\delta < 1.25^{3}$\\
  \hline 
  Eigen \cite{eigen2014depth} Coarse & GT Depth & - & 80m & 0.214 & 1.605 & 6.563 & 0.292 & 0.673 & 0.884 & 0.957\\ 
  Eigen \cite{eigen2014depth} Fine & GT Depth & - & 80m & 0.203 & 1.548 & 6.307 & 0.282 & 0.702 & 0.890 & 0.958\\ 
  Liu \cite{liu2015learning} & GT Depth & - & 80m & 0.201 & 1.584 & 6.471 & 0.273 & 0.68 & 0.898 & 0.967\\
  Zhou \cite{zhou2017unsupervised} & - & - & 80m & 0.208 & 1.768 & 6.856 & 0.283 & 0.678 & 0.885 & 0.957 \\
  Yang \cite{Yang2017unsupervised} &- &- & 80m &0.182 &1.481 &6.501 &0.267 &0.725 &0.906 &0.963 \\
  Vid2Depth \cite{mahjourian2018unsupervised} & - & - & 80m & 0.163 & 1.240 & 6.220 & 0.250 & 0.762 & 0.916 & 0.968 \\ 
  LEGO \cite{yang2018lego} &- &Yes & 80m &0.162 &1.352 &6.276 &0.252 &0.783 &0.921 &0.969 \\
  GeoNet \cite{yin2018geonet} &-  &Yes & 80m  &0.155 &1.296 &5.857 &0.233 &0.793 &0.931 &0.973 \\
  DDVO \cite{wang2018learning} &- &- & 80m &0.151 &1.257 &5.583 &0.228 &0.810 &0.936 &0.974 \\
  Godard \cite{godard2018digging}  &- &- & 80m  &0.133 &1.158 &5.370 &0.208 &0.841 &0.949 &0.978 \\
  Yang \cite{yang2018every}  &- &Yes & 80m  &0.131 &1.254 &6.117 &0.220 &0.826 &0.931 &0.973 \\
  
  \hline

  Ours (Baseline) & - & - & 80m   &0.1417    &1.1385    &5.5205    &0.2186    &0.8203    &0.9415    &0.9762  \\
  
  Ours (M) & - &Yes & 80m & 0.1412    & 1.0258    &5.2905   & 0.2153    &0.8160    &0.9452    & 0.9791 \\
    
  Ours (R) & - & - & 80m  &0.1231    &1.4367    &5.3099    &0.2043    & 0.8705    &0.9514    &0.9765 \\

  Ours (M+R) & - &Yes & 80m  & \textbf{0.1087}    & \textbf{0.8250}    &\textbf{4.7503}    & \textbf{0.1866}    & \textbf{0.8738}    & \textbf{0.9577}    & \textbf{0.9825}  \\
  \hline
  \end{tabular}
  }
  \caption{Evaluation of depth estimation of our method compared to the state-of-the-art. Separate results of the motion model (M), the online refinement one (R), and both (M+R) are presented. 
  For the purple columns, lower is better, for the yellow ones higher is better. KITTI dataset.}
    \label{tab:kitti_eigen}
\end{table*}

\begin{table*}[t]
  \centering
  \resizebox{1.0\textwidth}{!}{
  \begin{tabular}{|l|c|c|c||c|c|c|c|c|c|c|}
  \hline
  Method & Supervision? & Motion? & Cap & \cellcolor{col1}Abs Rel & \cellcolor{col1}Sq Rel & \cellcolor{col1}RMSE  & \cellcolor{col1}RMSE log & \cellcolor{col2}$\delta < 1.25 $ & \cellcolor{col2}$\delta < 1.25^{2}$ & \cellcolor{col2}$\delta < 1.25^{3}$\\
  \hline
  Garg \ea~\cite{garg2016unsupervised}* & - & - & 50m & 0.169 & 1.08 & 5.104 & 0.273 & 0.740 & 0.904  & 0.962 \\
  Mahjourian \ea~\cite{mahjourian2018unsupervised} & - & - & 50m & 0.155 & 0.927 & 4.549 & 0.231 & 0.781 & 0.931 & 0.975 \\
  GeoNet \cite{yin2018geonet} & - & - & 50m & 0.147 & 0.936 & 4.348 & 0.218 & 0.810 & 0.941 & 0.977 \\
  DDVO~\cite{wang2018learning}$\dagger$ & - & - & 50m & 0.1436 & 0.9348 & 4.2338 & 0.2144 & 0.8267 & 0.9446 & 0.9774 \\
  \hline
  Ours (Baseline) & - & - & 50m & 0.1343 & 0.8229 & 4.1078 & 0.2038 & 0.8365 & 0.9506 & 0.9802 \\
  Ours (R) & - & - & 50m   & 0.1141 & 0.9284 & 3.8777 & 0.1897 & 0.8841 & 0.9571 & 0.9792 \\
  Ours (M) & - &Yes & 50m  & 0.1350  &  0.7912  &  4.0573  &  0.2031  &  0.8311  &  0.9527 &   0.9822 \\
  Ours (M+R) & - &Yes & 50m & \textbf{0.1030} & \textbf{0.6217} & \textbf{3.5546} & \textbf{0.1749} & \textbf{0.8866} & \textbf{0.9632} & \textbf{0.9846} \\
  \hline
  \end{tabular}
  }
  \vspace{10pt}
  \caption{Evaluation of depth estimation of our method compared on the KITTI dataset, for 50 meters range cap. Methods marked with an asterik (*) use depth computed from disparities as ground truth, and are trained on stereo images. Results marked with $\dagger$ were computed by us using predictions that the authors provided.
}
\label{tab:kitti_eigen_50m}
\vspace{-10pt}
\end{table*}

\begin{figure*}
\begin{center}
  \includegraphics[width=0.245\linewidth]{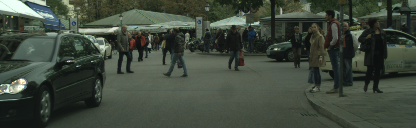}
  \includegraphics[width=0.245\linewidth]{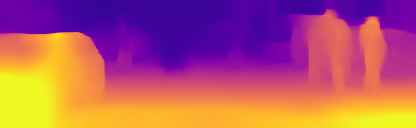}
  \includegraphics[width=0.245\linewidth]{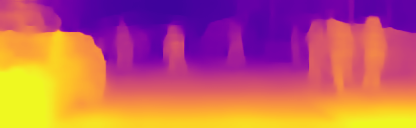}
  \includegraphics[width=0.245\linewidth]{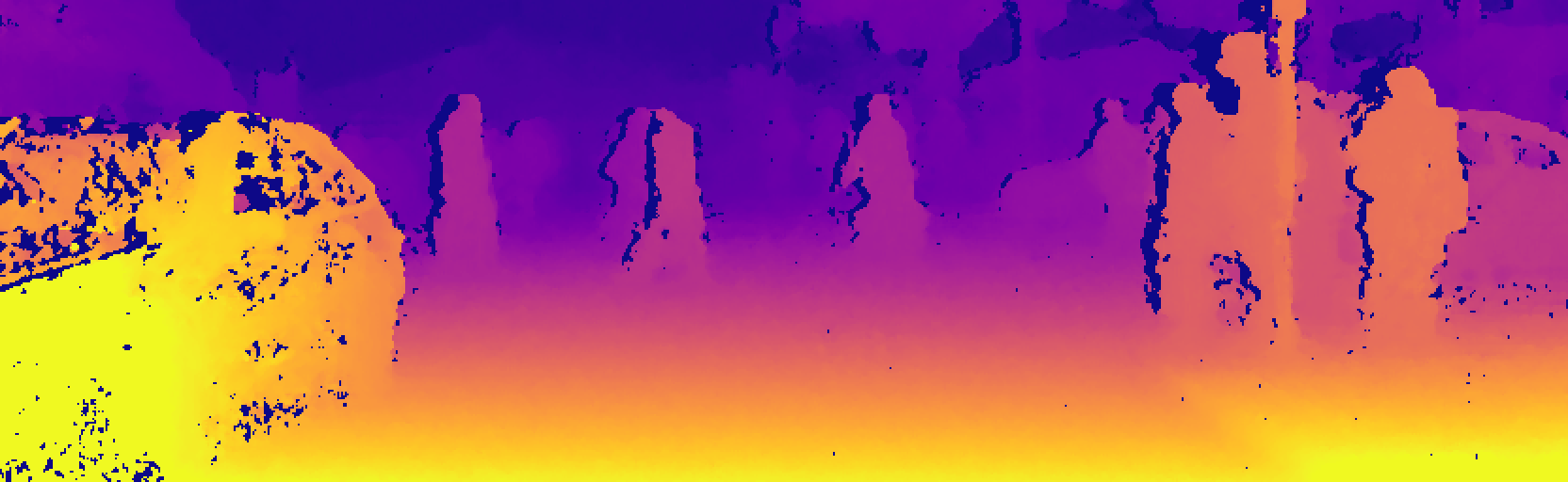}\\
  
  \includegraphics[width=0.245\linewidth]{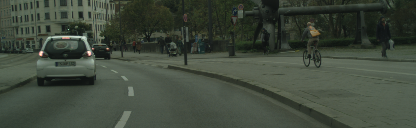}
  \includegraphics[width=0.245\linewidth]{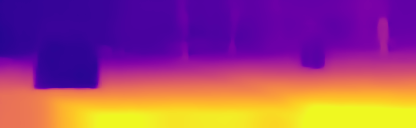}
  \includegraphics[width=0.245\linewidth]{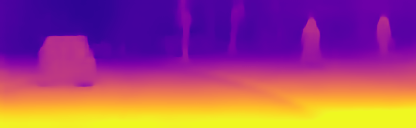}
  \includegraphics[width=0.245\linewidth]{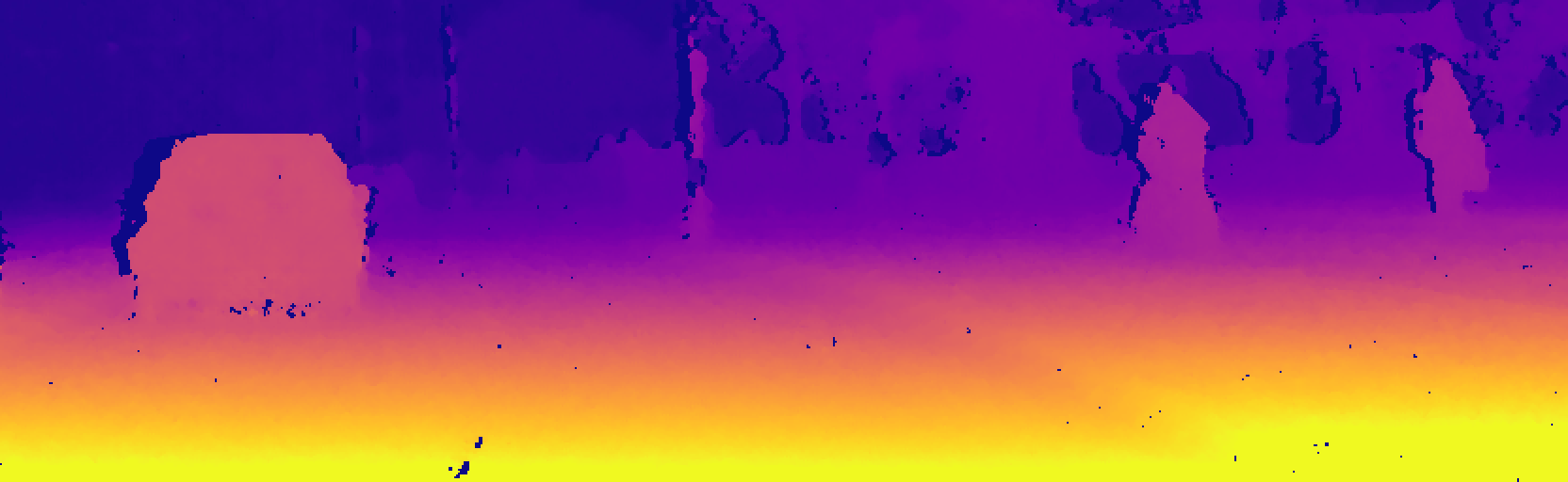}\\
  
  \includegraphics[width=0.245\linewidth]{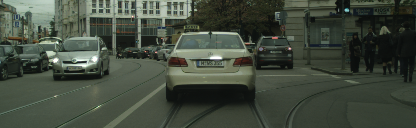}
  \includegraphics[width=0.245\linewidth]{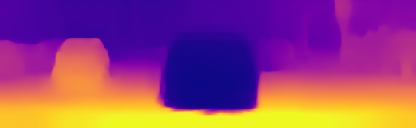}
  \includegraphics[width=0.245\linewidth]{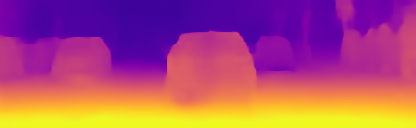}
  \includegraphics[width=0.245\linewidth]{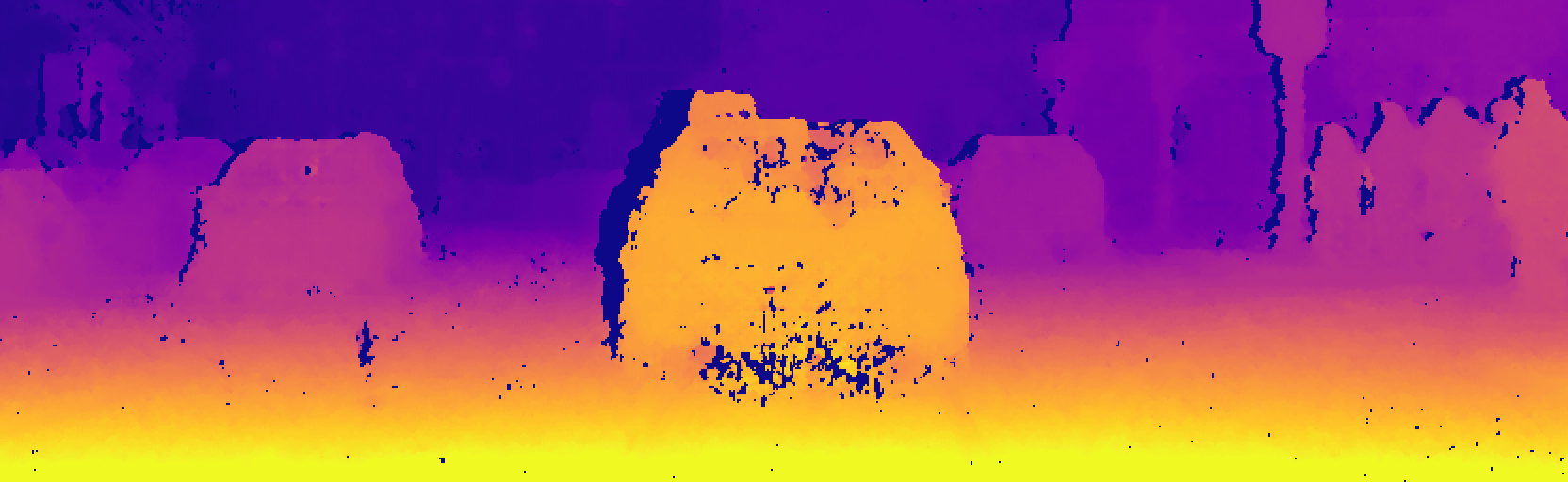}\\
  
  \includegraphics[width=0.245\linewidth]{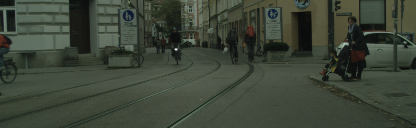}
  \includegraphics[width=0.245\linewidth]{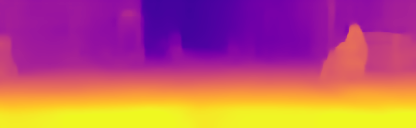}
  \includegraphics[width=0.245\linewidth]{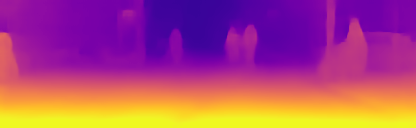}
  \includegraphics[width=0.245\linewidth]{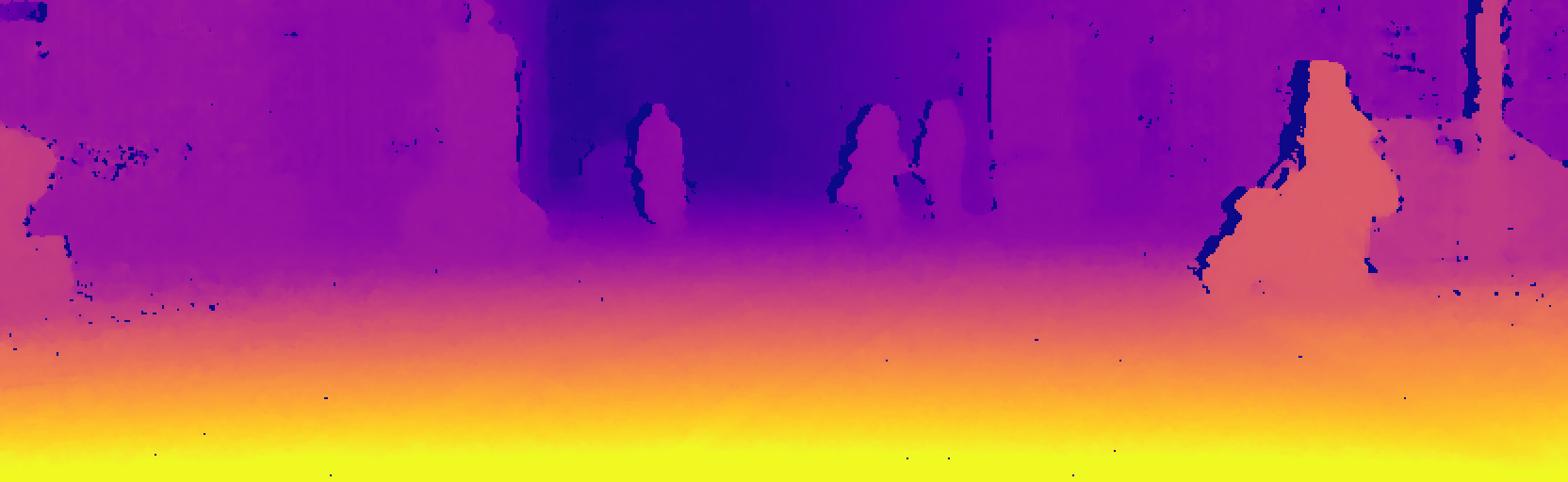}\\

\end{center}
\vspace{-4pt}
  \caption{Examples of depth estimation with the motion model (M) on highly dynamic scenes. City\-scapes dataset; from left to right: image input, baseline, ours, ground truth. A common failure case for dynamic scenes in monocular methods are objects moving with the camera itself. These objects are projected into infinite depth in prior work. Our method correctly estimates depth notably here, particularly on moving vehicles and persons.
  }
\label{fig:main_city}
\vspace{-4pt}
\end{figure*}

\begin{figure*}
\centering
\begin{tabular}{@{\hskip 1mm}c@{\hskip 1mm}c@{\hskip 1mm}c@{}}

\small{Input} & \small{Godard \ea~\cite{godard2018digging} Mono} & \small{GeoNet \cite{yin2018geonet}}\\

\includegraphics[width=0.33\linewidth]{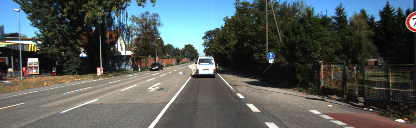} &
\includegraphics[width=0.33\linewidth]{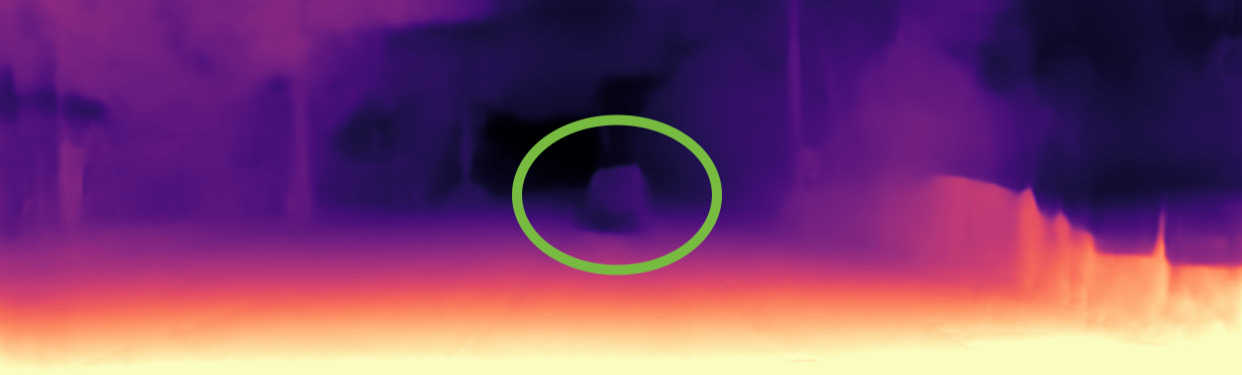} &
\includegraphics[width=0.33\linewidth]{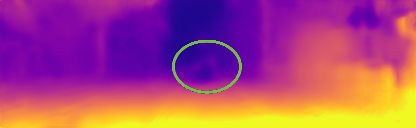}\\

\small{DDVO \cite{wang2018learning}} & \small{Baseline} & \small{Ours (M)}\\

\includegraphics[width=0.33\linewidth]{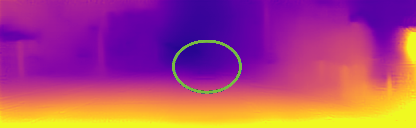} &
\includegraphics[width=0.33\linewidth]{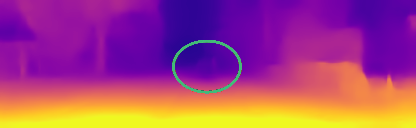} &  
\includegraphics[width=0.33\linewidth]{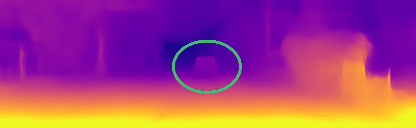} \\

\end{tabular}
   \caption{Example showing a common failure case for monocular methods, which is handled correctly by our motion model due to imposed size constraints. KITTI dataset.}
\label{fig:forward_motion2}
    \vspace{-10pt}
\end{figure*}

\subsection{Imposing Object Size Constraints}
Previous work has pointed out a significant limitation for monocular methods \cite{godard2018digging} \cite{yang2018every} \cite{wang2018learning} - that cars moving in front at roughly the same speed often get projected into infinite depth e.g. \cite{godard2018digging,yang2018every}. This is because the object in front shows no apparent motion, and if the network estimates it as being infinitely far away, the reprojection error is almost reduced to zero which is preferred to the correct case.
Previously, only methods with stereo images as input were able to solve this problem. Instead, we propose a different approach. Since the main problem stems from the fact that if the model has no knowledge about object scales, it could explain the same object motion by placing an object very far away and predicting large motion, or placing it close and predicting small motion, we here let the model learn the scales of objects as part of the training process. Assuming a weak prior on the height of certain objects, we can get an approximate depth estimation for it given its segmentation mask and the camera intrinsics using $D_{\text{approx}}(p;h)\approx f_y \frac{p}{h}$ where $f_y\in \mathbb{R}$ is the focal length, $p\in \mathbb{R}$ our height prior in world units, and $h\in \mathbb{N}$ the height of the respective segmentation blob in pixels. A loss term on the scale of each object $i$ ($i=1\dots N$) is added to the main loss. Let $t(i): \mathbb{N}\rightarrow \mathbb{N}$ define a category ID for any object $i$, and $p_j$ be a learnable height prior for each category ID $j$. Let $D$ be a depth map estimation and $S$ the corresponding object outline mask per object $O_i$ ($\odot$ is the element-wise multiplication). Then the loss is:

$$L_{sc}=\sum_{i=1}^{N} \Vert \frac{D \odot O_i(S)}{\overline{D}}-\frac{D_{\text{approx}}(p_{t(i)}; h(O_i(S)))}{\overline{D}}\Vert$$

\noindent We scale by $\overline{D}$, which is the mean estimated depth of the middle frame, to reduce a potential issue of trivial loss reduction by jointly shrinking priors and the depth prediction range.
To our knowledge this is the first method to address common degenerative cases in a fully monocular training setup in 3D.

In addition to the above-mentioned losses, the full loss includes the photometric reconstruction loss, the SSIM loss, a depth smoothness loss~\cite{zhou2017unsupervised,wang2018learning}. The loss is also applied on $4$ image resolutions.

\begin{table*}[t]
  \centering
  \resizebox{0.9\textwidth}{!}{
  \begin{tabular}{|l|c|c|c|c|c|c|c|c|c|}
  \hline
  Method &Train &Test & \cellcolor{col1}Abs Rel & \cellcolor{col1}Sq Rel & \cellcolor{col1}RMSE  & \cellcolor{col1}RMSE log & \cellcolor{col2}$\delta < 1.25 $ & \cellcolor{col2}$\delta < 1.25^{2}$ & \cellcolor{col2}$\delta < 1.25^{3}$\\
  \hline 
  Godard \ea\cite{godard2018digging}* & C  & K   &0.233 &3.533 &7.412 &0.292 &0.700 &0.892 &0.953\\
  \hline
  Ours (R) & C & K & 0.1696  &  1.7083  &  6.0151  &  0.2412  &  0.7840  &  0.9279  &  0.9703\\
  Ours (M) & C & K & 0.1876  &  1.3541  &  6.3166  &  0.2641  &  0.7135  &  0.9046  &  0.9667 \\
  Ours (M+R) & C & K & \textbf{0.1529}  &  \textbf{1.1087}  &  \textbf{5.5573}  &  \textbf{0.2272}  &  \textbf{0.7956}  &  \textbf{0.9338}  &  \textbf{0.9752}  \\
  \hline
  Pilzer \ea\cite{pilzer2018unsupervised}* &C  &C  &0.440 &6.036 &5.443 &0.398 &0.730 &0.887 &0.944 \\
  Pilzer \ea\cite{pilzer2018unsupervised}* &C  &C  &0.467 &7.399 &5.741 &0.493 &0.735 &0.890 &0.945 \\
  
  \hline
  Ours (R) & C & C & 0.2218  &  5.7374  &  8.6133  &  0.2584  &  0.7738  &  0.9076  &  0.9542\\
  Ours (M) & C & C &  \textbf{0.1454}  &  \textbf{1.7368}  &  7.2798  &  0.2046  &  0.8130  &  \textbf{0.9415}  &  \textbf{0.9775}\\
  Ours (M+R) & C & C & 0.1511  &  2.4916  &  \textbf{7.0237}  &  \textbf{0.2023}  &  \textbf{0.8255}  &  0.9372  &  0.9721\\
  \hline
  \end{tabular}
  }
  \vspace{5pt}
  \caption{Depth prediction results when training on City\-scapes. Evaluation on both KITTI (K) and City\-scapes (C) is shown here; 80m range cap. Methods marked with an asterik (*) might use a different cropping as the exact parameters were not available. 
  }
    \label{tab:main_city}
    \vspace{-10pt}
\end{table*}

\subsection{Test Time Refinement Model}
With the above-mentioned model, depth can be predicted from a single, still image during inference. If multi-frames are available during inference too, one may take advantage of that and learn on the fly.
More specifically, we propose to further optimize the model weights during inference  which allows the model to {\it adapt} to the environment online.
Thus, the model will be training for a number of steps, while performing inference. In doing that, we also show that even with very limited temporal resolution (i.e., three-frame sequences), the quality of depth predictions, both qualitatively and quantitatively, improves.

\begin{figure} [h!]
    \centering
    \includegraphics[width=1.0\linewidth]{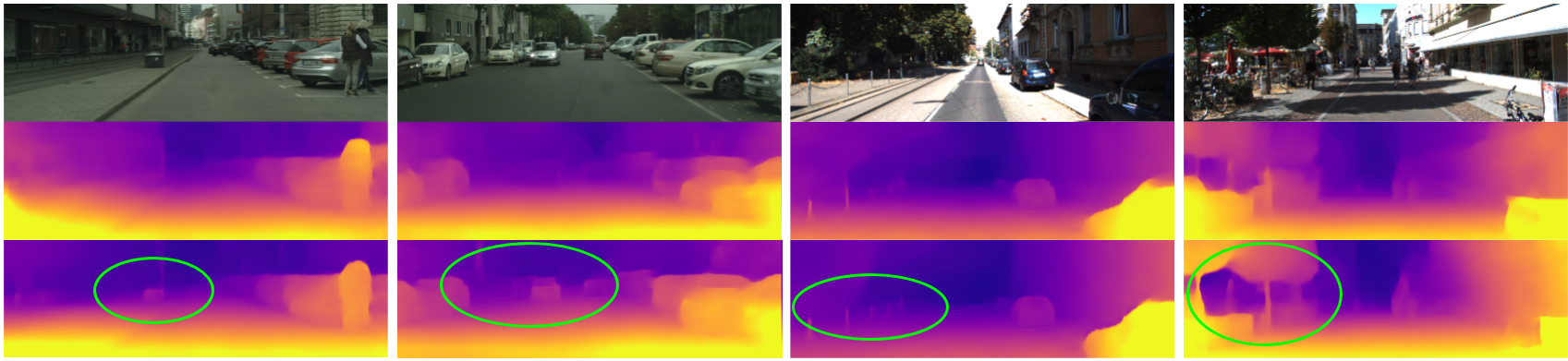}
    \caption{Online refinement model (R). Cityscapes (left columns), KITTI (right columns). The model is trained on KITTI.
    }
    \label{fig:kitti_refinement}
\end{figure}

\section{Experimental Results}
We conduct experiments on depth estimation, ego-motion estimation and on transfer learning to new environments, using common metrics and protocols for evaluation adopted by prior methods. We report results on the two main benchmarks for depth and ego-motion evaluation: the KITTI dataset~\cite{geiger2013vision} and the Cityscapes dataset~\cite{cordts2016cityscapes}.

\textbf{Results on the KITTI Dataset.}
Figure~\ref{fig:main_kitti} visualizes the results of our method compared to the ground truth provided by a sensor and Tables~\ref{tab:kitti_eigen} and~\ref{tab:kitti_eigen_50m} show quantitative results. Improvement over the baseline and over previous methods in the literature is observed. Our method outperforms competitive models that use motion~\cite{yang2018every} and~\cite{yin2018geonet}. 
Furthermore, our results which are trained in monocular setup, are close to methods which use stereo or a combination of stereo and monocular, e.g.~\cite{godard2017monodepth,yang2018every,godard2018digging}. More results can be seen in~\cite{casser2019depth}.

\textbf{Experimental Results on the Cityscapes Dataset.}
In this section we evaluate our method on the Cityscapes dataset, whhich contains many dynamic scenes. Table~\ref{tab:main_city} shows our experimental results when training on the Cityscapes data, and then evaluating on KITTI (without fine-tuning). We also show evaluation on the Cityscapes dataset itself, which contains many moving objects. These experiments clearly demonstrate the benefit of our method for dynamic scenes as we see significant improvements in depth estimation. We observe that the improvements are due to the appropriate depth learning of many moving objects (Figure~\ref{fig:main_city}) enabled by the motion model. We further note that these are new results and training and {\it testing} on Cityscapes is not customarily done, as seen in the table, since the dataset is very challenging.

\textbf{Motion Model.}
We here further examine the effects of the motion model. Figure~\ref{fig:main_city} shows several examples of dynamic scenes from the Cityscapes dataset, which contain many moving objects. We note that our baseline, which is by itself a top performer on KITTI, is failing on moving objects. Our method makes a notable difference both qualitatively (Figure~\ref{fig:main_city}) and quantitatively (see Table~\ref{tab:main_city}).   
Figure~\ref{fig:forward_motion2} further compares our results with previous monocular methods in the case of a moving vehicle in front of the ego-motion vehicle. As seen our approach is the only one that can extract its depth.
Another benefit provided by the motion model is that it learns to predict motions of individual objects in 3D, which can be available for inference if an object mask is specified~\cite{casser2019depth}. In the general case, object masks are not needed for depth or ego-motion inference.

\begin{table} [h!]
\centering
\resizebox{0.45\textwidth}{!}{
\begin{tabular}{|l|c|c|}
\hline
Method & Seq. $09$  & Seq. $10$ \\
\hline
Mean Odometry  & 0.032 $\pm 0.026$  & $0.028 \pm 0.023$ \\
ORB-SLAM (short)  & $0.064 \pm 0.141$ & $0.064 \pm 0.130$ \\
Vid2Depth (Mahjourian 2018) & $0.013 \pm 0.010$ & $0.012 \pm 0.011$ \\
Godard (Godard 2018)$\dagger$  & $0.023 \pm 0.013$ & $0.018 \pm 0.014$ \\
Zhou (Zhou 2017)$\dagger$   & $0.021 \pm 0.017$  & $0.020 \pm 0.015$ \\
GeoNet (Yin 2018)  & $ 0.012 \pm 0.007$ & $0.012 \pm 0.009$ \\
ORB-SLAM (full)*  & $0.014 \pm  0.008$ & $0.012 \pm 0.011$ \\
Ours   & $\mathbf{0.011 \pm 0.006}$    & $\mathbf{0.011 \pm 0.010}$\\
\hline
\end{tabular}
}
\caption{Quantitative evaluation of odometry on the KITTI Odometry test sequences. Methods using more information than a set of rolling 3-frames are marked (*). Models that are trained on a different part of the dataset are marked ($\dagger$).}
\label{fig:kitti_vo}
\end{table}

\begin{figure*}[t]
    \centering        
    \includegraphics[width=0.98\linewidth]{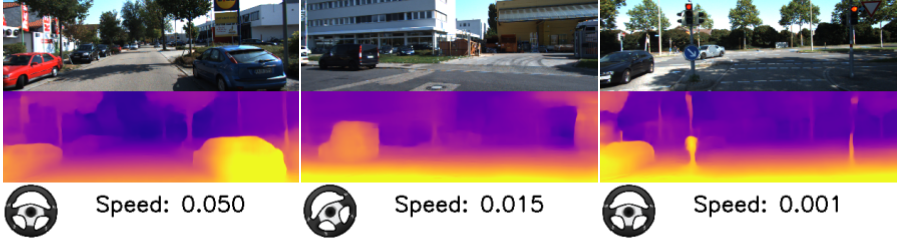}
    \caption{Ego-motion results visualized as vehicle speed (in non-metric network units) and turn indicator at the bottom: driving forward (left), slowing down and taking a turn (middle), stopping for a red light (right).}
    \label{fig:ego-res}
\end{figure*}

\begin{figure}
    \begin{center}
    \includegraphics[width=1.0\linewidth]{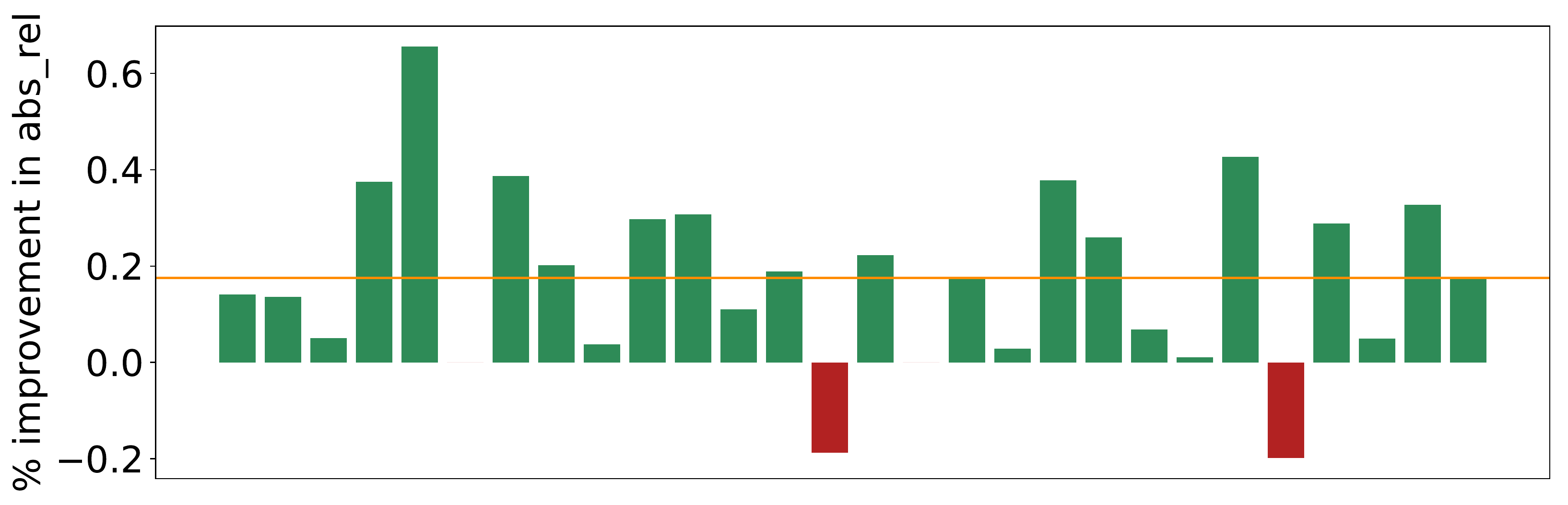}
    \end{center}
    \caption{Relative improvements achieved by the online refinement model per frame. As seen, most frames benefit from online refinement. The red horizontal line marks the mean improvement. KITTI dataset. }
    \label{fig:kitti_fusion_delta}
\end{figure}

\textbf{Refinement Model.}
Figure~\ref{fig:kitti_refinement} shows results of the refinement method only.
We can see improvements of the refinement model on both KITTI and Cityscapes datasets for a model trained on KITTI.
As seen for both evaluating on KITTI or Cityscapes dataset the refinement is helpful in recovering the geometry structure better.
Of note that in the case of Cityscape (left), this is testing across datasets.
Figure~\ref{fig:kitti_fusion_delta} further shows improvements per frame by the online refinement model. As seen, most frames benefit from refinement and improve their depth estimation. More online refinement results, demonstrated on an indoor dataset, collected by the Fetch robot are shown in~\cite{casser2019depth}.

\textbf{Visual Odometry Results.}
The ego-motion results are shown in Table~\ref{fig:kitti_vo}. The experiments are conducted by a standard protocol adopted by prior work~\cite{zhou2017unsupervised,godard2018digging} on the KITTI odometry dataset. As seen our algorithm outperforms state-of-the-art methods, even the ones that use more temporal information.  Handling of motion is the biggest contributing  factor to improving the ego-motion estimation of our algorithm. Figure~\ref{fig:ego-res} shows results on the KITTI sequence.

\textbf{Experiments Discussion.}
As shown previously (in Tables~\ref{tab:kitti_eigen}, ~\ref{tab:kitti_eigen_50m}, and~\ref{tab:main_city}) our method benefits from both motion and online refinement, but each component works in different extents.
For example, the motion net although extremely beneficial for City\-scapes, a dataset with many dynamic scenes, affects the metrics to moderate amounts in KITTI, which reflects the scarcity of motion in this dataset. Online refinement, on the other hand, is generally useful, but some confusion may arise when applying it solo on a data with a lot of motion, e.g., City\-scapes. When both online refinement and motion are applied we have much better results than the baselines.

\begin{figure}
    \begin{center}
    \includegraphics[width=1.0\linewidth]{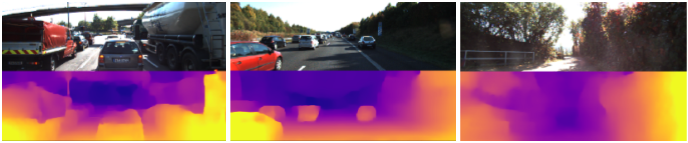}
    \end{center}
    \caption{Example failure cases of our model can be due to unfamiliar objects, for example, the large tank truck and the bridge in the right image, or side vegetation (in the middle) and fence (right image).}
    \label{fig:result_failures}
\end{figure}

Figure~\ref{fig:result_failures} shows several typical examples of failures for depth estimation for the motion-only model. These can generally happen in scenes which are considerably different than scenes seen during training, for example scenes in which the average depth is very low over the full image, or images with new objects, such as a tank truck or a bridge.  

\textbf{Notes on the Evaluation Procedure and Revised Results.}
As pointed out by Godard \ea~\cite{godard2018digging}, the evaluation code released by \cite{zhou2017unsupervised} contains an inaccuracy where the depth ground truth on the KITTI dataset is computed with respect to the camera instead of the LIDAR. Fortunately, in practice, the effects of this are rather subtle as the displacement is not very large. We know that at least the results of Zhou \ea~\cite{zhou2017unsupervised}, Mahjourian \ea~\cite{mahjourian2018unsupervised}, MonoDepth~\cite{godard2017monodepth}, Pilzer \ea~\cite{pilzer2018unsupervised} and GeoNet~\cite{yin2018geonet} are affected, as they adopted the same evaluation code. To be able to better compare to these methods, all numbers reported in the main paper use the old evaluation code for both caps at $50m$ and $80m$. We show results using the revised evaluation code in Table~\ref{tab:kitti_eigenfixed}. For all methods where we have raw predictions available, we recompute their scores and also include them in the table.

\begin{table*}[t]
  \centering
  \resizebox{0.9\textwidth}{!}{
  \begin{tabular}{|l|c|c|c||c|c|c|c|c|c|c|}
  \hline
  Method & Supervision? & Motion? & Cap & \cellcolor{col1}Abs Rel & \cellcolor{col1}Sq Rel & \cellcolor{col1}RMSE  & \cellcolor{col1}RMSE log & \cellcolor{col2}$\delta < 1.25 $ & \cellcolor{col2}$\delta < 1.25^{2}$ & \cellcolor{col2}$\delta < 1.25^{3}$\\
  
  \hline 
  Mahjourian \ea~\cite{mahjourian2018unsupervised}$\dagger$ & - & - & 50m & 0.1563 & 0.9602 & 4.4051 & 0.2320 & 0.7940 & 0.9323 & 0.9732\\
  GeoNet~\cite{yin2018geonet}$\dagger$ & - & - & 50m & 0.1512 & 0.9455 & 4.3627 & 0.2224  & 0.8023 & 0.9387 & 0.9759\\
  DDVO~\cite{wang2018learning}$\dagger$ & - & - & 50m & 0.1488 & 0.9498 & 4.2551 & 0.2195 & 0.8181 & 0.9423 & 0.9765 \\
  
  \hline
  Ours (Baseline) & - & - & 50m & 0.1388 & 0.8344 & 4.1306 & 0.2087 & 0.8291 & 0.9480 & 0.9793 \\
  Ours (M) & - &Yes & 50m & 0.1377 & 0.7918 & 4.0514 & 0.2066 & 0.8260 & 0.9506 & 0.9814 \\
  Ours (R) & - & - & 50m  & 0.1175 & 0.9398 & 3.8833 & 0.1932 & 0.8799 & 0.9560 & 0.9787 \\
  Ours (M+R) & - &Yes & 50m  & \textbf{0.1052} & \textbf{0.6231} & \textbf{3.5591} & \textbf{0.1779} & \textbf{0.8834} & \textbf{0.9620} & \textbf{0.9841}  \\

  \hline
  \hline

  Train set mean & - & - & 80m & 0.361 & 4.826 & 8.102 & 0.377 & 0.638 & 0.804 & 0.894 \\
  Mahjourian \ea~\cite{mahjourian2018unsupervised}$\dagger$ & - & - & 80m & 0.1635 & 1.2467 & 5.9579 & 0.2478 & 0.7752 & 0.9197 & 0.9680\\
  GeoNet~\cite{yin2018geonet}$\dagger$ & - & - & 80m & 0.1590 & 1.3032 & 5.8732 & 0.2379 & 0.7853 & 0.9286 & 0.9713 \\
  DDVO~\cite{wang2018learning}$\dagger$ & - & - & 80m & 0.1562 & 1.2750 & 5.6136 & 0.2336 & 0.8015 & 0.9330 & 0.9724\\
  Godard \ea~\cite{godard2018digging} & - & - & 80m & 0.137 & 1.153 & 5.353 & 0.212 & 0.836 & 0.947 & 0.978 \\
  \hline
  Ours (Baseline) & - & - & 80m & 0.1463 & 1.1506 & 5.5520 & 0.2236 & 0.8127 & 0.9386 & 0.9751 \\
  Ours (M) & - &Yes & 80m & 0.1439 & 1.0247 & 5.2914 & 0.2189 & 0.8110 & 0.9429 & 0.9782 \\
  Ours (R) & - & - & 80m  & 0.1265 & 1.4453 & 5.3122 & 0.2078 & 0.8663 & 0.9501 & 0.9759 \\
  Ours (M+R) & - &Yes & 80m  & \textbf{0.1108} & \textbf{0.8254} & \textbf{4.7619} & \textbf{0.1897} & \textbf{0.8704} & \textbf{0.9563} & \textbf{0.9819}  \\
  
  \hline
  \end{tabular}
  }
  \vspace{10pt}
  \caption{Evaluation of depth estimation of our method using the \textit{revised} evaluation code on KITTI, testing individual contributions of motion and online refinement components. We re-evaluate related methods if predictions are available, as marked with $\dagger$. As before, our method outperforms every competing one.}
  \label{tab:kitti_eigenfixed}
    \vspace{-10pt}
\end{table*}

\section{Conclusions}
This paper addresses the monocular depth and ego-motion problem by modeling individual objects' motion in 3D, and an online refinement algorithm which is beneficial for transfering learned models to new environments. The algorithm allows application to videos with dynamic scenes and motion. Results on two major and challenging benchmarks datasets, KITTI and City\-scapes, for depth and ego-motion prediction are presented. We also showed successful transfer across datasets.

\textbf{Acknowledgments.}
We would like to thank Ayzaan Wahid for helping us with data collection, members of the Brain team for discussions, and
Chaoyang Wang, Zhenheng Yang, Zhichao Yin and Jianping Shi for their generous sharing of results. We also would like to thank Cl\'ement Godard for helping with reproducing some previous results.

{\small
\bibliographystyle{ieee_fullname}
\bibliography{references}

\begin{thebibliography}{10}\itemsep=-1pt

\bibitem{casser2019depth}
Vincent Casser, Soeren Pirk, Reza Mahjourian, and Anelia Angelova.
\newblock Depth prediction without the sensors: Leveraging structure for
  unsupervised learning from monocular videos.
\newblock In {\em Thirty-Third AAAI Conference on Artificial Intelligence
  (AAAI-19)}, 2019.

\bibitem{cordts2016cityscapes}
M. Cordts, M. Omran, S. Ramos, T. Rehfeld, M. Enzweiler, R. Benenson, U.
  Franke, S. Roth, and B. Schiele.
\newblock The cityscapes dataset for semantic urban scene understanding.
\newblock In {\em CVPR}, 2016.

\bibitem{eigen2014depth}
D. Eigen, C. Puhrsch, and R. Fergus.
\newblock Depth map prediction from a single image using a multi-scale deep
  network.
\newblock {\em NIPS}, 2014.

\bibitem{garg2016unsupervised}
Ravi Garg, Gustavo Carneiro, and Ian Reid.
\newblock Unsupervised cnn for single view depth estimation: Geometry to the
  rescue.
\newblock {\em ECCV}, 2016.

\bibitem{geiger2013vision}
A. Geiger, P. Lenz, C. Stiller, and R. Urtasun.
\newblock Vision meets robotics: The kitti dataset.
\newblock {\em The International Journal of Robotics Research},
  32(11):1231--1237, 2013.

\bibitem{godard2018digging}
C. Godard, O.~Mac Aodha, and G. Brostow.
\newblock Digging into self-supervised monocular depth estimation.
\newblock {\em arxiv.org/pdf/1806.01260}, 2018.

\bibitem{godard2017monodepth}
Clement Godard, Oisin~Mac Aodha, and Gabriel~J. Brostow.
\newblock Unsupervised monocular depth estimation with left-right consistency.
\newblock {\em CVPR}, 2017.

\bibitem{he2017mask}
K. He, G. Gkioxari, P. Doll{\'a}r, and R. Girshick.
\newblock Mask r-cnn.
\newblock In {\em ICCV}, 2017.

\bibitem{Kuznietsov2017semisupervised}
Y. Kuznietsov, J. Stuckler, and B Leibe.
\newblock Sfm-net: Learning of structure and motion from video.
\newblock {\em CVPR}, 2017.

\bibitem{laina2016deeper}
Iro Laina, Christian Rupprecht, Vasileios Belagiannis, Federico Tombari, and
  Nassir Navab.
\newblock Deeper depth prediction with fully convolutional residual networks.
\newblock {\em arXiv:1606.00373}, 2016.

\bibitem{liu2015learning}
F. Liu, C. Shen, G. Lin, and I. Reid.
\newblock Learning depth from single monocular images using deep convolutional
  neural fields.
\newblock {\em PAMI}, 2015.

\bibitem{mahjourian2017geometry}
Reza Mahjourian, Martin Wicke, and Anelia Angelova.
\newblock Geometry-based next frame prediction from monocular video.
\newblock {\em Intelligent Vehicles Symposium}, 2017.

\bibitem{mahjourian2018unsupervised}
Reza Mahjourian, Martin Wicke, and Anelia Angelova.
\newblock Unsupervised learning of depth and ego-motion from monocular video
  using 3d geometric constraints.
\newblock In {\em Proceedings of the IEEE Conference on Computer Vision and
  Pattern Recognition}, pages 5667--5675, 2018.

\bibitem{pilzer2018unsupervised}
A. Pilzer, D. Xu, M. Puscas, E. Ricci, and N. Sebe.
\newblock Unsupervised adversarial depth estimation using cycled generative
  networks.
\newblock {\em 3DV}, 2018.

\bibitem{ummenhofer2017demon}
Benjamin Ummenhofer, Huizhong Zhou, Jonas Uhrig, Nikolaus Mayer, Eddy Ilg,
  Alexey Dosovitskiy, and Thomas Brox.
\newblock Demon: Depth and motion network for learning monocular stereo.
\newblock {\em CVPR}, 2017.

\bibitem{wang2018learning}
C. Wang, J. Buenaposada, R. Zhu, and S. Lucey.
\newblock Learning depth from monocular videos using direct methods.
\newblock {\em CVPR}, 2018.

\bibitem{yang2018every}
Z. Yang, P. Wang, Y. Wang, W. Xu, and R. Nevatia.
\newblock Every pixel counts: Unsupervised geometry learning with holistic 3d
  motion understanding.
\newblock {\em ECCV Workshop}, 2018.

\bibitem{yang2018lego}
Zhenheng Yang, Peng Wang, Yang Wang, Wei Xu, and Ram Nevatia.
\newblock Lego: Learning edge with geometry all at once by watching videos.
\newblock {\em CVPR}, 2018.

\bibitem{Yang2017unsupervised}
Z. Yang, P. Wang, W. Xu, L. Zhao, and R. Nevatia.
\newblock Unsupervised learning of geometry with edge-aware depth-normal
  consistency.
\newblock {\em arXiv:1711.03665}, 2017.

\bibitem{yin2018geonet}
Shi~J. Yin, Z.
\newblock Geonet: Unsupervised learning of dense depth, optical flow and camera
  pose.
\newblock {\em CVPR}, 2018.

\bibitem{zhan2018unsupervised}
H. Zhan, R. Garg, C.S. Weerasekera, K. Li, H. Agarwal, and I. Reid.
\newblock Unsupervised learning of monocular depth estimation and visual
  odometry with deep feature reconstruction.
\newblock {\em CVPR}, 2018.

\bibitem{zhou2017unsupervised}
T. Zhou, M. Brown, N. Snavely, and D. Lowe.
\newblock Unsupervised learning of depth and ego-motion from video.
\newblock {\em CVPR}, 2017.

\end{thebibliography}
}

\end{document}